\documentclass[english,10pt, a4paper, twocolumn,utf8]{article}
\usepackage[T1]{fontenc}
\usepackage[utf8]{inputenc}
\usepackage{babel}
\usepackage{array}
\usepackage{verbatim}
\usepackage{textcomp}
\usepackage{graphicx}
\usepackage[unicode=true,
 bookmarks=true,bookmarksnumbered=true,bookmarksopen=true,bookmarksopenlevel=2,
 breaklinks=false,pdfborder={0 0 1},backref=false,colorlinks=false]
 {hyperref}
\hypersetup{pdftitle={Improving Chemical Autoencoder Latent Space and Molecular De-novo Generation Diversity with Heteroencoder},
 pdfauthor={Esben Jannik Bjerrum},
 pdfsubject={De novo generation of molecules with heteroencoders (autoencoders)},
 pdfkeywords={Deep Learning, RNN, LSTM, de novo molecule design, molecular autoencoders, molecular heteroencoders}}

\makeatletter

\providecommand{\tabularnewline}{\\}

\usepackage[]{babel} 
 \usepackage{cite}

\usepackage{amsmath,amsfonts,amsthm} 

\usepackage[svgnames]{xcolor} 

\usepackage[small, labelfont=bf, up, textfont=it]{caption} 

\usepackage{booktabs} 

\usepackage{lastpage} 

\usepackage{graphicx} 

\usepackage{enumitem} 
\setlist{noitemsep} 

\usepackage{sectsty} 
\allsectionsfont{\usefont{OT1}{phv}{b}{n}} 

\usepackage{mdframed} 

\usepackage[font=small,labelfont=bf]{caption}
\makeatletter
\g@addto@macro\@floatboxreset\centering
\makeatother

\usepackage{geometry} 

\usepackage[T1]{fontenc} 

\usepackage{XCharter} 

\usepackage{fancyhdr} 
\newcommand{\headrulecolor}[1]{\patchcmd{\headrule}{\hrule}{\color{#1}\hrule}{}{}}
\newcommand{\footrulecolor}[1]{\patchcmd{\footrule}{\hrule}{\color{#1}\hrule}{}{}}

\renewcommand{\footrulewidth}{1pt} 

\headrulecolor{DarkGoldenrod}
\footrulecolor{DarkGoldenrod}
\fancypagestyle{firstpage}{ 
\renewcommand{\footrulewidth}{0.0pt} 
}

\newcommand{\authorstyle}[1]{{\large\usefont{OT1}{phv}{b}{n}\color{DarkRed}#1}} 

\newcommand{\institution}[1]{{\footnotesize\usefont{OT1}{phv}{m}{sl}\color{Black}#1}} 

\usepackage{titling} 

\pretitle{
\date{}
\vspace{-70pt} 
\vspace{35pt} 
\fontsize{18}{24}\usefont{OT1}{phv}{b}{n}\selectfont 
\color{DarkRed} 
}

\posttitle{\par\vskip 15pt} 

\preauthor{} 

\postauthor{
\vspace{-40pt} 
}

\usepackage{lettrine} 
\usepackage{fix-cm}	

\newcommand{\initial}[1]{ 
\lettrine[lines=3,findent=4pt,nindent=0pt]{
\color{DarkGoldenrod}
{#1}
}{}%
}

\usepackage{xstring} 

\newcommand{\lettrineabstract}[1]{

\mdframed[backgroundcolor=gray!20,hidealllines=true]
\vspace{5pt} 
\StrLeft{#1}{1}[\firstletter] 
\initial{\firstletter}\textbf{\StrGobbleLeft{#1}{1}} 
\vspace{5pt} 
\endmdframed 

\vspace{10pt} 
}

\author{
\authorstyle{Esben Jannik Bjerrum\textsuperscript{1,*}, Boris Sattarov\textsuperscript{2}} 
\newline\newline 
\textsuperscript{1}\institution{Wildcard Pharmaceutical Consulting, Zeaborg Science Center, Frødings Allé 41, 2860 Søborg, Denmark.}\\ 
\textsuperscript{2}\institution{Science Data Software LLC, 14914 Bradwill Court, Rockville, Maryland 20850, United States.}\\ 
\textsuperscript{*}\institution{Corresponding Author: \href{mailto://esben@wildcardconsulting.dk}{esben@wildcardconsulting.dk}} 
}

\makeatother

\begin{document}

\twocolumn[   \begin{@twocolumnfalse}

\title{Improving Chemical Autoencoder Latent Space and Molecular \textit{De-novo}
Generation Diversity with Heteroencoders }

\maketitle
\thispagestyle{fancy}
\renewcommand{\footrulewidth}{0.0pt}
\lhead{}
\chead{}
\rhead{}
\lettrineabstract{Chemical autoencoders are attractive models as they combine chemical space navigation with possibilities for de-novo molecule generation in areas of interest. This enables them to produce focused chemical libraries around a single lead compound for employment early in a drug discovery project. Here it is shown that the choice of chemical representation, such as SMILES strings, has a large influence on the properties of the latent space. It is further explored to what extent translating between different chemical representations influences the latent space similarity to the SMILES strings or circular fingerprints. By employing SMILES enumeration for either the encoder or decoder, it is found that the decoder has the largest influence on the properties of the latent space. Training a sequence to sequence heteroencoder based on recurrent neural networks(RNNs) with long short-term memory cells (LSTM) to predict different enumerated SMILES strings from the same canonical SMILES string gives the largest similarity between latent space distance and molecular similarity measured as circular fingerprints similarity. Using the output from the bottleneck in QSAR modelling of five molecular datasets shows that heteroencoder derived vectors markedly outperforms autoencoder derived vectors as well as models built using ECFP4 fingerprints, underlining the increased chemical relevance of the latent space. However, the use of enumeration during training of the decoder leads to a markedly increase in the rate of decoding to a different molecules than encoded, a tendency that can be counteracted with more complex network architectures.}

 \end{@twocolumnfalse} ]

\section*{Introduction}

Autoencoders have emerged as deep learning solutions to turn molecules
into latent vector representations as well as decode and sample areas
of the latent vector space\cite{Gomez-Bombarelli2016,Blaschke2017_autoencoders,Xu2017_seq2seq}.
An autoencoders consists of an encoder which compresses and changes
the input information into a bottleneck layer and a decoder part which
recreates the original input from the compressed vector representation
(the latent space vector). After training, the autoencoder can be
reassembled into the encoder which can be used to calculate vector
representations of the molecules. These can be used as a sort of molecular
fingerprints or GPS for the chemical space of the molecules. The decoder
can be used to translate back from the latent representation to the
molecular representation used during training, such as simplified
molecular-input line-entry system (SMILES). This makes it possible
to use the decoder as a steered solution for molecular \emph{de-novo}
generation, as the probability outputs of the decoder can be sampled,
creating molecules which are novel but close to the point in latent
space. Alternatively, the molecules of the nearby latent space can
be explored by randomly permuting the latent vector.
\begin{figure}
\caption{\label{fig:Can2can-Enumeration-challenge}Enumeration challenge of
a sequence to sequence model trained on canonical SMILES. The non-canonical
SMILES of the same molecule is projected to different parts of the
latent space reduced to two dimensions with principal components analysis
(PCA). The small blue dots are the test set used for fitting the PCA.
Some clustering of the enumerated SMILES can be observed.}

\includegraphics[width=1\columnwidth]{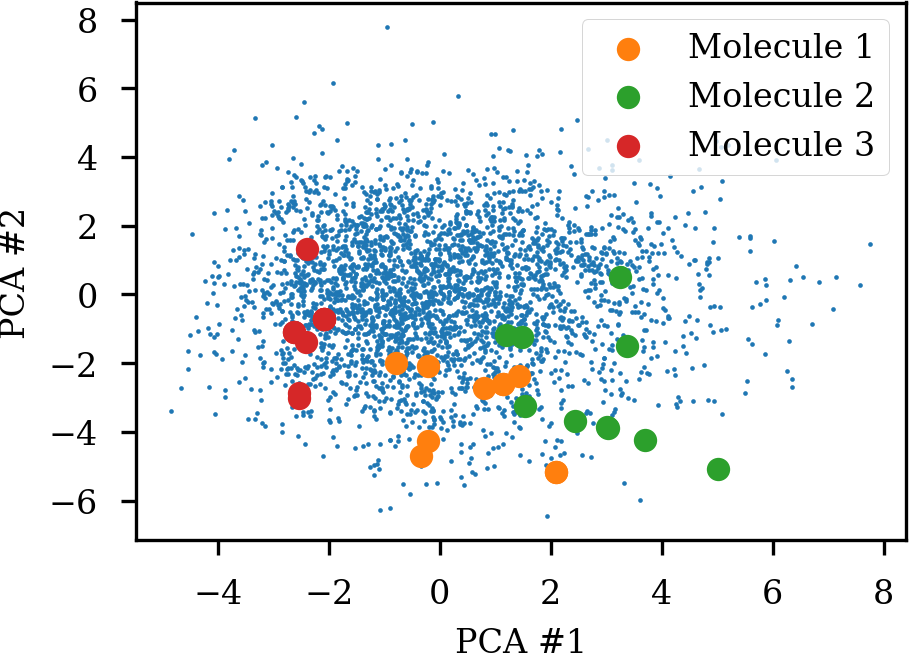}
\end{figure}

Various encoder-decoder architectures have been proposed as well as
the latent space has been regularized and manipulated using variational
autoencoders\cite{Gomez-Bombarelli2016} and adversarial autoencoders\cite{Blaschke2017_autoencoders}.
Both convolutional neural networks (CNNs), as well as recurrent neural
networks (RNNs) have been used for the encoder part\cite{Gomez-Bombarelli2016,Xu2017_seq2seq,Blaschke2017_autoencoders},
whereas the decoder part has mostly been based on RNNs with either
gated recurrent units (GRU)\cite{Chung2014} or long short-term memory
cells (LSTM)\cite{Hochreiter1997} to enable longer range sequence
memory.

A famous painting of René Magritte, ``The Treachery of Images'',
shows a pipe, and also has the text ``Ceci n'est pas une pipe'':
This is not a pipe. The sentence is true as it is a painting of pipe,
not the pipe itself, kindly reminding us that representation is not
reality. Autoencoders based on SMILES strings\cite{Weininger1970}
face the same fundamental issue. Is the latent space a representation
of the molecules or is it a condensed representation of the SMILES
strings representing the molecules?

Due to serialization of the molecular graph, a single molecule has
multiple possible SMILES strings, which has been exploited as data
augmentation with the SMILES enumeration technique\cite{Bjerrum2017}.
A simple challenge with different SMILES representations of the same
molecule suggest that the latent space is more related to the SMILES
string than to the molecule, which has also previously been noted\cite{Blaschke2017_autoencoders}.
Figure \ref{fig:Can2can-Enumeration-challenge} shows the same three
molecules after projecting into the latent space of an RNN to RNN
autoencoder. The molecules end up being projected to very different
areas of the latent space, although some clustering can be observed.
The latent space is thus likely a mixture of SMILES representation
information and chemical information representation. One way of solving
this challenge could be to use special engineered networks and graph
based approaches\cite{Li2018} for molecular generation.

\begin{figure*}
\caption{\label{fig:Concept_illustratoin}Chemical heteroencoders are similar
to autoencoders but translates from one format of the information
or representation of the information to the other. The molecule toluene
can be represented as a canonical SMILES strings, in different enumerated
SMILES or via a 2D embedding. The autoencoder converts the canonical
SMILES string to the latent space and back again (blue arrow), whereas
many more possibilities exists for heteroencoders (green arrows).}

\includegraphics[width=1\textwidth]{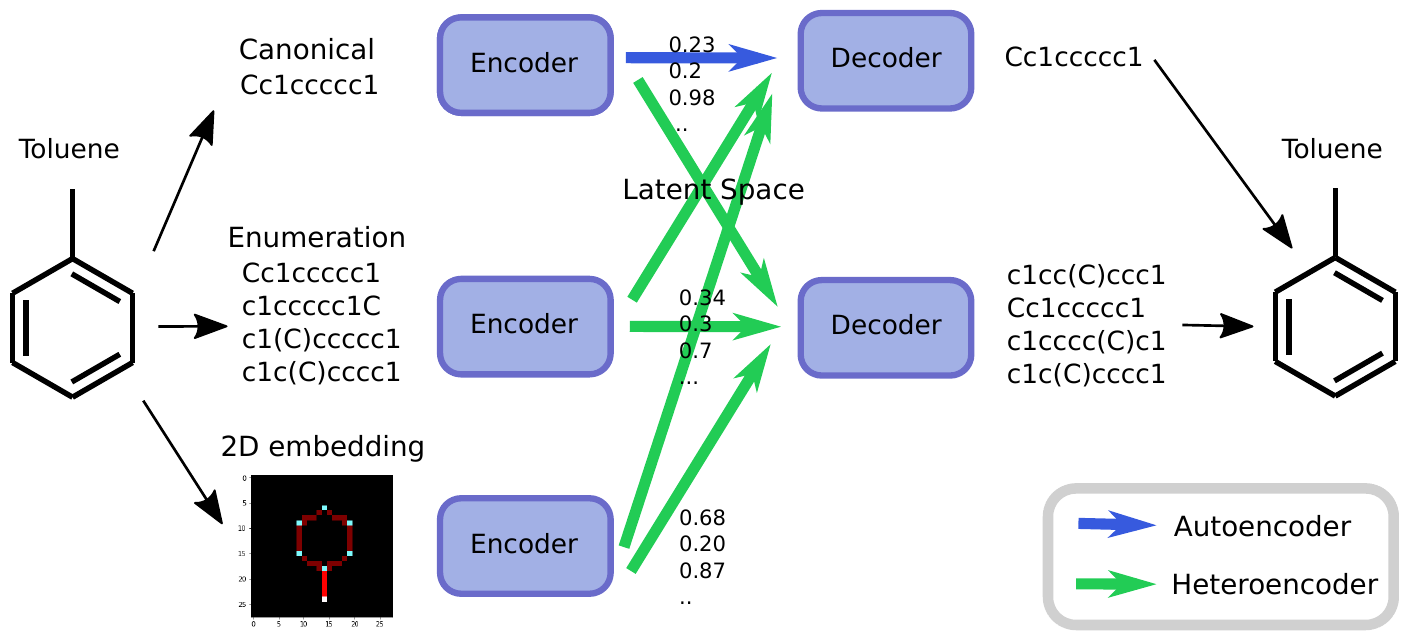}
\end{figure*}

As an alternative to engineering the outcome, it is here suggested
that it is possible to use SMILES enumeration or chemception image
embedding\cite{goh2017chemception} to create chemical heteroencoders.
The concept is illustrated in Figure \ref{fig:Concept_illustratoin}.
 By translating from one format or representation of the molecule
to the other, the encoder-decoder network is forced to identify the
latent information behind both representations. This should in principle
lead to a more chemically relevant latent space, independent of the
formats or canonicalization used. 

Here, the choice of representation and enumeration is explored for
both training the encoder or decoder and the latent space similarity
to SMILES and scaffold based metrics calculated. Moreover, it is tested
if these changes influence the properties of the decoder when used
for \emph{de-novo} design of molecules. Further, an optimized and
expanded heteroencoder architectures trained on ChEMBL23 datasets
are used to extract latent vectors for subsequent use as input to
QSAR models of five different molecular datasets.

\section*{Methods}

\subsection*{Datasets}

\subsubsection*{GDB-8}

The GDB-8 dataset\cite{Polishchuk2013,Ruddigkeit2012} was downloaded
and split randomly into a train and test set using a 0.9 to 0.1 ratio.

\subsubsection*{ChEMBL23}

Structures were extracted from the ChEMBL23 database\cite{Gaulton2017}
and validated using in-house rules at Science Data Software LLC (salts
were stripped, solvents removed, charges neutralized and stereo information
removed). Maximum available length of the canonical SMILES string
allowed for a molecule was 100 characters. 10 thousand molecules were
selected randomly for the held out test set. From the remainder of
the 1.2 million molecules was randomly selected a training set of
400 thousand molecules and a validation set of 300 thousand molecules
for used during training procedures.

\subsubsection*{QSAR datasets}

Five experimental datasets were used, spanning physico-chemical properties
as well as bioactivity. Four datasets (IGC50, BCF, MP, LD50) for QSAR
modeling were downloaded from the EPA Toxicity Estimation Software
Tool\cite{EpiSuiteDownload} webpage\cite{EpiSuiteSoftware} and used
as is without any additional standardization. The solubility was obtained
from the supplementary information of \cite{huuskonen2000estimation}.
Information of the datasets are shown in Table \ref{tab:QsarDatasetInfo}.
\begin{table*}
\caption{\label{tab:QsarDatasetInfo}The datasets used for QSAR modelling.}

\begin{tabular}{c>{\centering}p{6cm}>{\centering}p{2.3cm}>{\centering}p{2.3cm}}
\hline 
Label & Endpoint & Endpoint values span & Number of Molecules\tabularnewline
\hline 
BCF & Bioconcentration factor, the logarithm of the ratio of the concentration
in biota to its concentration in the surrounding medium (water)\cite{Arnot2006} & -1.7 to 5.7 & 541\tabularnewline
IGC50 & Tetrahymena pyriformis 50\% growth inhibition concentration (g/L)\cite{Schultz1997} & 0.3 to 6.4 & 1434\tabularnewline
LD50 & Lethal Dosis 50\% rats (mg/kg body weight)\cite{ChemIDplus} %
{}  & 0.5 to 7.1 & 5931\tabularnewline
MP & Melting point of solids at normal atmospheric pressure\cite{EpiSuiteDownload} & -196 to 493 & 7509\tabularnewline
Solubility & log water solubility (mol/L)\cite{huuskonen2000estimation} & -11.6 to 1.6 & 1297\tabularnewline
\hline 
\end{tabular}
\end{table*}

Datasets from EPA’s TEST suite were already split into train/test
sets in a 75/25\% ratio and were used accordingly. Molecules for the
solubility dataset were obtained by resolving CAS numbers from the
supporting info\cite{huuskonen2000estimation} and the dataset was
randomly split using the same ratio as the other QSAR datasets.

\subsubsection*{1D and 2D Vectorization}

SMILES were enumerated and vectorized with one-hot encoding as previously
described\cite{Bjerrum2017}. 2D vectorization was done similar to
the vectorization used in Chemception networks\cite{goh2017chemception}
with the following modifications: A PCA with three principal components
was calculated on atomic properties from the mendeleev python package\cite{mendeleev2014}
(dipole\_polarizability, electron\_affinity, electronegativity, vdw\_radius,
atomic\_volume, softness and hardness). The PCA scores were normalized
with min-max scaling to be between zero and one to create the atom
type encoding. PCA and scaling were performed with the Scikit-Learn
python package\cite{scikit-learn}. RDKit\cite{Landrum2016} was used
to compute 2D coordinates and extract information about atom type
and bond order. The normalized PCA scores of the atom types were used
to encode the first three layers and the bond order was used to encode
the forth layer. A fifth layer was used to encode the RDKit aromaticity
perception. The 2D coordinates of the RDKit molecule were rotated
randomly up to +/- 180\textdegree{} around the center of coordinates
before discretization into numpy\cite{walt2011numpy} floating point
arrays.

\subsubsection*{Neural Network Modeling for GDB-8 Dataset}

Sequence to Sequence RNN models were constructed using Keras v. 2.1.1\cite{chollet2015}
and Tensorflow v. 1.4\cite{Abadi2016b}. The first layer consisted
of 64 LSTM cells\cite{Hochreiter1997} used in batch mode. The final
internal memory (C) and hidden (H) states were concatenated and used
as input to a dense layer (the bottleneck) of 64 neurons with the
rectified linear unit activation function (ReLU)\cite{Nair2010}.
Two separate dense layers with ReLU activation functions were used
to decode the bottleneck outputs into the initial C and H states for
the RNN based decoder. The decoder consisted of a single layer of
64 LSTM cells trained with teacher forcing\cite{Williams1989} in
batch mode. The output from the LSTM cells was connected to a Dense
layer with a softmax activation function matching the dimensions of
the character set. A two-layer model was also constructed by increasing
the number of LSTM cells to 128 and the number of LSTM layers to two
in both the encoder and decoder. Accordingly, four separate dense
networks were used to decode the bottleneck layer into the initial
C and H states for the two LSTM layers in the decoder.

The networks were trained with mini-batches of 256 sequences for 300
epochs using the categorical cross entropy loss function and the Adam
optimizer\cite{Kingma2014} with an initial learning rate of 0.05.
The two layer model was trained with an initial learning rate of 0.01.
The loss was followed on the test set and the learning rate lowered
by a factor of two when no improvement in the test set loss had been
observed for 5 epochs.

After training in batch mode, three models were created from the parts
of the full model. A decoder model from the initial input to the output
of the bottleneck layer. A model to calculate the initial states of
the LSTM cells in the decoder, given the output of the bottleneck.
Lastly, a stateful decoder model was constructed by creating a model
with the exact same architecture as the decoder in the full model,
except the LSTM cells were used in stateful mode and the input vector
reduced to a size of one in the sequence dimension. After creation
of the stateful model, the weights for the networks were copied from
corresponding parts of the trained full model.

The image to sequence model CNN encoder was built from three different
Inception-like modules\cite{Szegedy2014}. The first module consisted
of a tower with a 1x1 2D convolutional layer (Conv2D) followed by
a 3x3 Conv2D, a tower with a 1x1 Conv2D layer followed by a 5x5 Conv2D
layer and a tower with just a single 1x1 Conv2D layer. The outputs
from the towers were concatenated and sent to the next module.

The standard inception module was constructed with a tower of 1x1
Conv2D layer followed by a 3x3 Conv2D layer, a tower with a 1x1 Conv2D
layer followed by a 5x5 Conv2D layer, an extra tower of a 1x1 Conv2D
layer followed by a 7x7 Conv2D but with only half the number of kernels
and a tower with a 3x3 Maxpooling layer followed by a 1x1 Conv2D layer.
All strides were 1x1. The outputs from the four towers were concatenated
and sent to the next module.

The inception reduction modules were similar to the standard module,
except they had no 7x7 tower and used a stride of 2x2.

A standard inception module was stacked with a reduction inception
module three times, giving 7 inception modules in total including
the initial one. The number of kernels was set to 32.

The outputs from the last inception module were flattened and followed
by a dropout layer with a dropout rate of 0.2. Lastly the output was
connected to the bottleneck consisting of a dense layer with the ReLU
activation function. The decoder part was constructed similar to the
sequence to sequence models described above with one layer LSTM cells.
The image to sequence model was trained similar to the sequence to
sequence models for 200 epochs.

The models are named after the training data in a encoder2decoder
naming scheme. ``Can'' is training data with canonical SMILES, where
``Enum'' designates that the input or output was enumerated during
the training. ``Img'' shows that the data was the 2D image embedding.

\subsubsection*{Similarity metrics}

SMILES strings sequence similarities were calculated as the alignment
score reported by the pairwise global alignment algorithm of the Biopython
package\cite{Cock2009_biopython}. The match score was set to 1, the
mismatch to -1, the gap opening to -0.5 and the gap extension to -0.05.
The fingerprint similarity metric was calculated on basis of circular
Morgan fingerprints with a radius of 2 as implemented in the RDKit
library\cite{Landrum2016}. The fingerprints were hashed to 2048 bits
and the similarity calculated with the RDKit packages FingerprintSimilarity
function. The latent space similarity between two molecules was calculated
as the negative logarithm to the Euclidean distance of the vector
coordinates.

\subsubsection*{Enumeration Challenge}

The encoder was used to calculate the latent space of the test set,
followed by a dimensionality reduction with standard principal components
analysis (PCA) as implemented in the Scikit-Learn package\cite{scikit-learn}.
Three molecules were converted to different SMILES strings with SMILES
enumeration\cite{Bjerrum2017}. The latent space coordinates of the
non-canonical SMILES were calculated with the encoder and transformed
and projected onto the visualization of the principal components from
the PCA analysis.

\subsubsection*{Error analysis of output}

The percentage of invalid SMILES was quantified as the number of produced
SMILES which could not be validated as molecules by RDKit. Subsequently
the equality of the input and output RDKit molecules was checked.
The similarity of the scaffold was checked by comparing the generalized
murcko scaffolds\cite{Bemis1996} including side-chains. The atom
equivalence was checked by comparing the molecular sum formulas. The
number and nature of bonds was compared via a ``bond sum formula''
by counting the number of single, double, triple and aromatic bonds.

\subsubsection*{Multinomial sampling of decoder}

Multinomial sampling was implemented as previously described\cite{Bjerrum2017_molgen}.
The sampling temperature was kept at 1.0.

\subsubsection*{Neural Network Modelling for the ChEMBL dataset}

The sequence to sequence autoencoder used for encoding the ChEMBL
data and extraction of vectors for QSAR modelling was programmed in
Python 3.6.1\cite{van1995python} using Keras version 2.1.5\cite{chollet2015}
with the tensorflow backend\cite{Abadi2016b}. The encoder consisted
of two bidirectional layers of 128 CuDNNLSTM cells in each one-way
layer. The final C and H states were concatenated and passed as input
to a dense layer with 256 neurons using the ReLU\cite{Nair2010} activation
function (the bottleneck). The output from the dense layer were decoded
by four parallel dense layers with the ReLU activation function, whose
outputs were used to set the initial C and H states of the decoder
LSTM layers. The decoder itself consisted of two unidirectional layers
of 256 CuDNNLSTM cells each . The decoder was trained under teacher
forcing as described for the simpler networks above. Every non-linear
activation was followed by Batch Normalization. No additional regularization
was used. 400k random structures from the CheMBL23 training set were
pre-enumerated 50-times for each SMILES string. The new 20M pairs
were shuffled and used in both a canonical to enumerated and an enumerated
to canonical setting and trained until model convergence. The same
400k canonical SMILES were also used to train an auto encoder from
canonical to canonical SMILES. For the enumerated to enumerated training
setting 50 pairs (when possible) of different SMILES strings were
created for each molecule of the training set. The network was trained
using mini-batches of 256 one-hot encoded SMILES strings, using the
Adam optimizer with an initial learning rate of 0.005. The training
was monitored and controlled by three callbacks. One callback monitored
the loss of the validation set and lowered the learning rate by a
factor two when no improvement had been observed for 2 epochs (ReduceLROnPlateu).
Another Callback stopped training when no improvement in the validation
set loss had been observed for 5 epochs (EarlyStopping), and the last
callback saved the model if the validation loss has improved (CheckPoint).
Models typically converged after approximately 40 epochs, which usually
took about 6 hours on a NVIDIA GTX 1080 Ti equipped server.

\subsubsection*{QSAR modelling}

Subsequent QSAR modelling was performed using the machine learning
capabilities of the Open Science Data Repository\cite{OpenDataRepository}.
An initial search for hyper parameters were performed after converting
the molecules into ECPF4 fingerprints (radius 2, 1024 bits). The hyper
parameter search for a neural network was performed using Tree of
Parzen Estimators (TPE) algorithm\cite{Bergstra2011} as implemented
in Hyperopt\cite{Hyperopt} with the search space bounds listed in
Table \ref{tab:Hyper-parameter-search}
\begin{table}

\caption{\label{tab:Hyper-parameter-search}Hyper parameter search space using
the Tree of Parzen estimator method in Hyperopt.}

\begin{tabular}{>{\centering}p{3cm}>{\centering}p{3cm}}
\hline 
Hyper parameter & Search space\tabularnewline
\hline 
Input dropout & 0.0-0.95\tabularnewline
Units per layer & 2 - 1024\tabularnewline
Kernel regularizer (L2) & 0.000001 - 0.1\tabularnewline
Kernel constraint (maxnorm) & 0.5 - 6\tabularnewline
Kernel initializer & lecun\_uniform' 'glorot\_uniform', 'he\_uniform', 'lecun\_normal',
'glorot\_normal', 'he\_normal'\tabularnewline
Batch normalization & Yes (after each activation), No\tabularnewline
Activation function & ReLU, SeLU\tabularnewline
Dropout & 0.0 - 0.95\tabularnewline
Number of hidden layers & 1 - 6\tabularnewline
Learning rate & 0.00001 - 0.1\tabularnewline
Optimizer & Adam, Nadam, RMSprop, SGD\tabularnewline
\hline 
\end{tabular}
\end{table}
. The performance on each dataset was optimized using 3-fold cross
validation on the training set. The performance of the model with
the final hyperparameters were subsequently tested on the held-out
test set using an ensemble of 10 models build during 10-fold cross
validation during training. The auto-/heteroencoders trained on the
ChEMBL23 molecules were subsequently used to encode the QSAR datasets
into vectors using the output from the bottleneck layer. The vectors
were used as input to the QSAR models. The same hyper parameters were
used as identified for the ECFP4 based models, with no further attempt
to optimize the hyper parameters of the feed forward neural networks
using the auto-/heteroencoder extracted inputs.

\section*{Results}

\subsubsection*{GDB-8 dataset based models}

All models were trained to full convergence and obtained close train
and test losses, the later listed in Table \ref{tab:Model Properties}
\begin{table*}
\caption{\label{tab:Model Properties}Properties of the models trained on different
different input and output representations of the GDB-8 dataset. All
values were calculated using the test dataset.}

\begin{tabular*}{1\textwidth}{@{\extracolsep{\fill}}>{\centering}p{2.7cm}cc>{\centering}p{2.7cm}>{\centering}p{2.7cm}>{\centering}p{3cm}>{\centering}p{2.7cm}}
\hline 
Model &  & Loss & \% Malformed SMILES & \% wrong molecule & $R^{2}$ Fingerprint metric & $R^{2}$ Sequence metric\tabularnewline
\hline 
\hline 
Can2Can &  & 0.0005 & 0.1 & 0.0 & 0.24 & \textbf{0.58}\tabularnewline
Img2Can &  & 0.02 & 0.0 & 8.0 & 0.05 & 0.18\tabularnewline
Enum2Can &  & 0.03 & 1.0 & 17.1 & 0.37 & 0.53\tabularnewline
Can2Enum &  & 0.18 & 1.7 & 50.3 & \textbf{0.58} & 0.55\tabularnewline
Enum2Enum &  & 0.21 & 2.2 & 66.8 & 0.49 & 0.40\tabularnewline
Enum2Enum 2-layer &  & 0.13 & 0.3 & 14.7 & 0.45 & 0.55\tabularnewline
\hline 
 &  &  &  &  &  & \tabularnewline
\end{tabular*}
\end{table*}
. The models trained on enumerated SMILES output have a markedly larger
final loss, but all models show a low degree of malformed SMILES when
sampling the latent space vectors calculated from the test set.

\subsubsection*{Molecular and sequence similarity}

\begin{figure}
\caption{\label{fig:Examples-of-alignments}Examples of optimal SMILES alignments
of a molecule with two other molecules. The score is +1 for character
match, -1 for mismatch, gap openings -0.5 and gap extension -0.05.
Gaps are show with dashes, ``-'', and are not SMILES single bonds.}

\includegraphics[width=1\columnwidth]{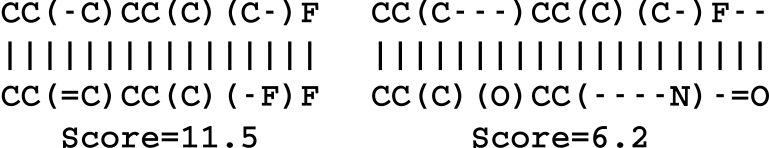}
\end{figure}

\begin{figure}
\caption{\label{fig:R2_seqsim}Scatter plot of the latent space similarities
and the alignment scores of the SMILES strings.}

\includegraphics[width=1\columnwidth]{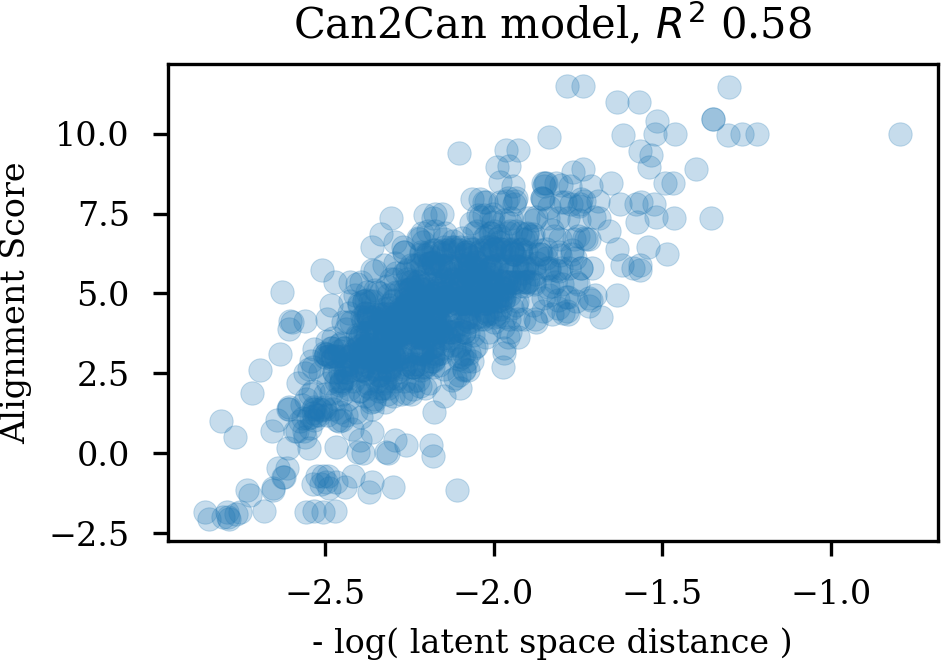}
\end{figure}

\begin{figure}
\caption{\label{fig:R2_morgansim}Scatter plot of the latent space similarities
and the circular fingerprint similarities.}

\includegraphics[width=1\columnwidth]{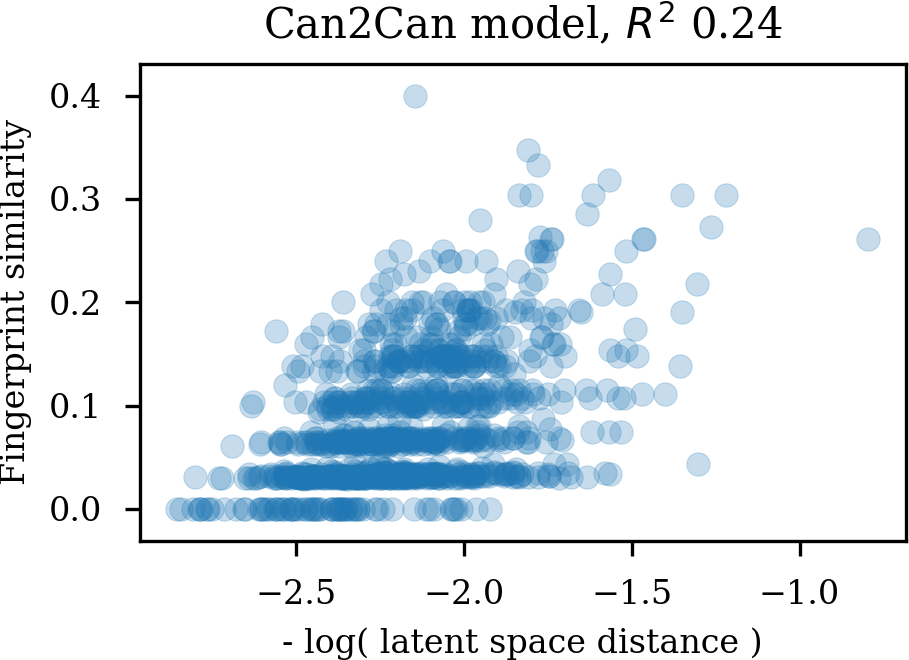}
\end{figure}

Using the same reference molecule, similarity metrics were calculated
based on the latent space vectors of the test set, Morgan fingerprints
and sequence alignment scores, followed by calculation of the correlation
coefficients ($R^{2}$). Examples of SMILES alignments are shown in
Figure \ref{fig:Examples-of-alignments}
 for two different alignments. Figure \ref{fig:R2_seqsim}
shows an example scatter plot of the alignment scores and latent space
similarity for the first molecule and the rest of the molecules in
the test set. The correlation between the same latent space similarity
measurement and the Morgan fingerprint similarity was intended as
a metric of the scaffold similarity independent of the SMILES strings,
and an example scatter plot is shown in Figure \ref{fig:R2_morgansim}.
 Both the sequence alignment score and the fingerprint based similarity
have correlation with the latent space similarity, which shows that
the latent space is at least somehow related to our traditional understanding
of similarities between molecules. The properties and the correlations
of all the models trained on the GDB-8 dataset are listed in Table
\ref{tab:Model Properties}. The models with a decoder trained on
canonical SMILES show a markedly larger correlation between the latent
space and the SMILES sequence similarity metric than between the fingerprint
based similarity and the latent space. In contrast, the fingerprint
and sequence similarities correlations to the latent space similarity
are more on the same level when the decoder is trained using enumerated
SMILES. The heteroencoder based on the image embedding of the molecule
has the lowest correlations, indicating a markedly different or noisy
latent space.

\begin{figure}
\caption{\label{fig:molsim_can2can}Molecules similar in latent space using
the can2can model. The reference molecule is in the upper left corner
and similarity drops row-wise in normal reading direction.}

\includegraphics[width=1\columnwidth]{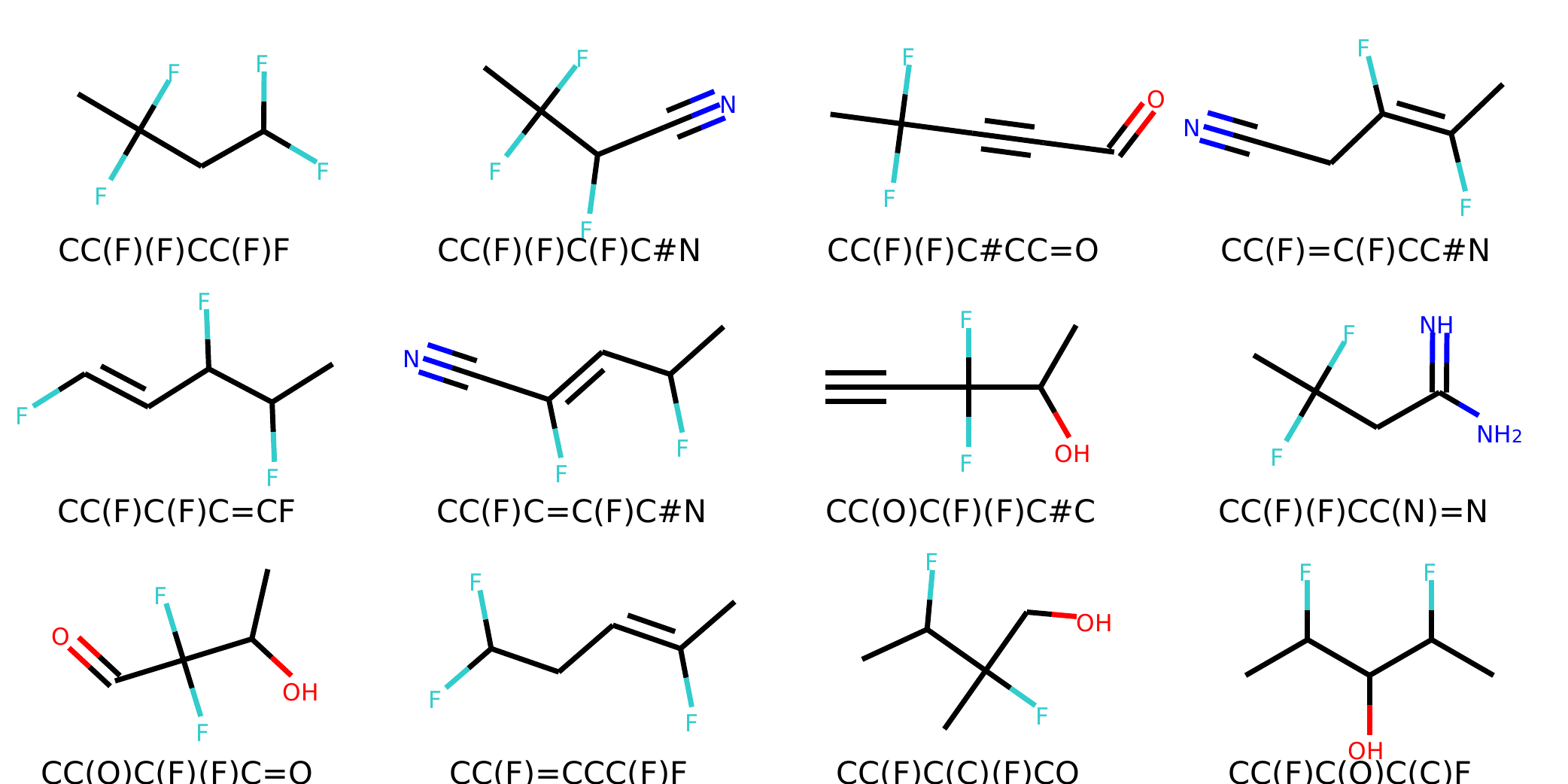}
\end{figure}

\begin{figure}
\caption{\label{fig:MolSim_can2enum}Molecules similar in latent space using
the can2enum model. The reference molecule is in the upper left corner
and similarity drops row-wise in normal reading direction.}

\includegraphics[width=1\columnwidth]{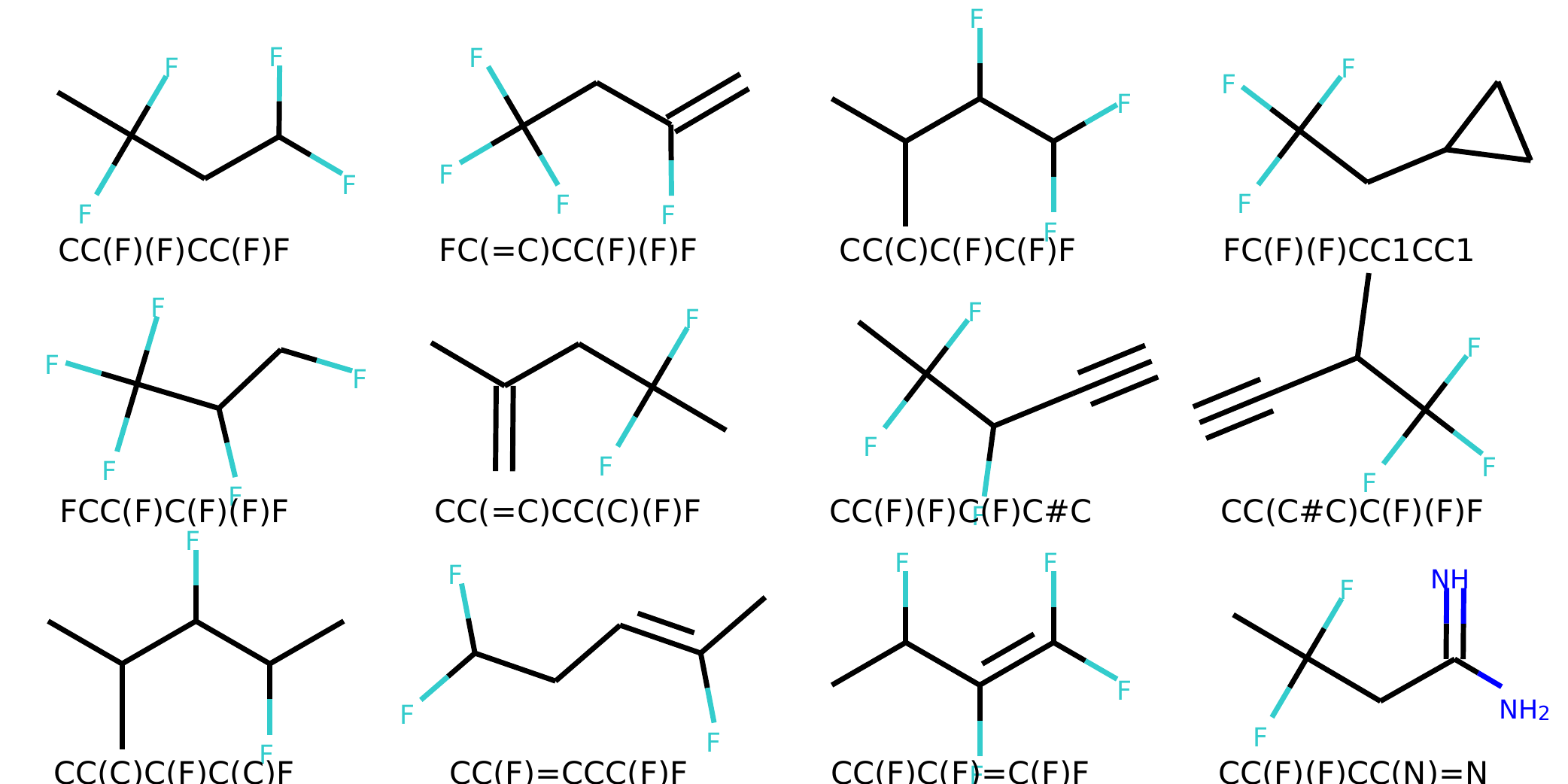}
\end{figure}

Figure \ref{fig:molsim_can2can} and \ref{fig:MolSim_can2enum} show
a result of similarity searching in the latent space of the test set
using a query molecule.
 The molecules from the can2enum model seems qualitatively more similar
than the ones that are most similar in the latent space produced by
the can2can model. There is overlap between the two sets, so in some
respects the two latent spaces are related.

\subsubsection*{Error analysis}

\begin{figure}
\caption{\label{fig:Venn-diagram-of}Venn diagram of the errors encounted during
molecule reconstruction of 1000 molecules for the GDB-8 can2enum model.}
\includegraphics[width=1\columnwidth]{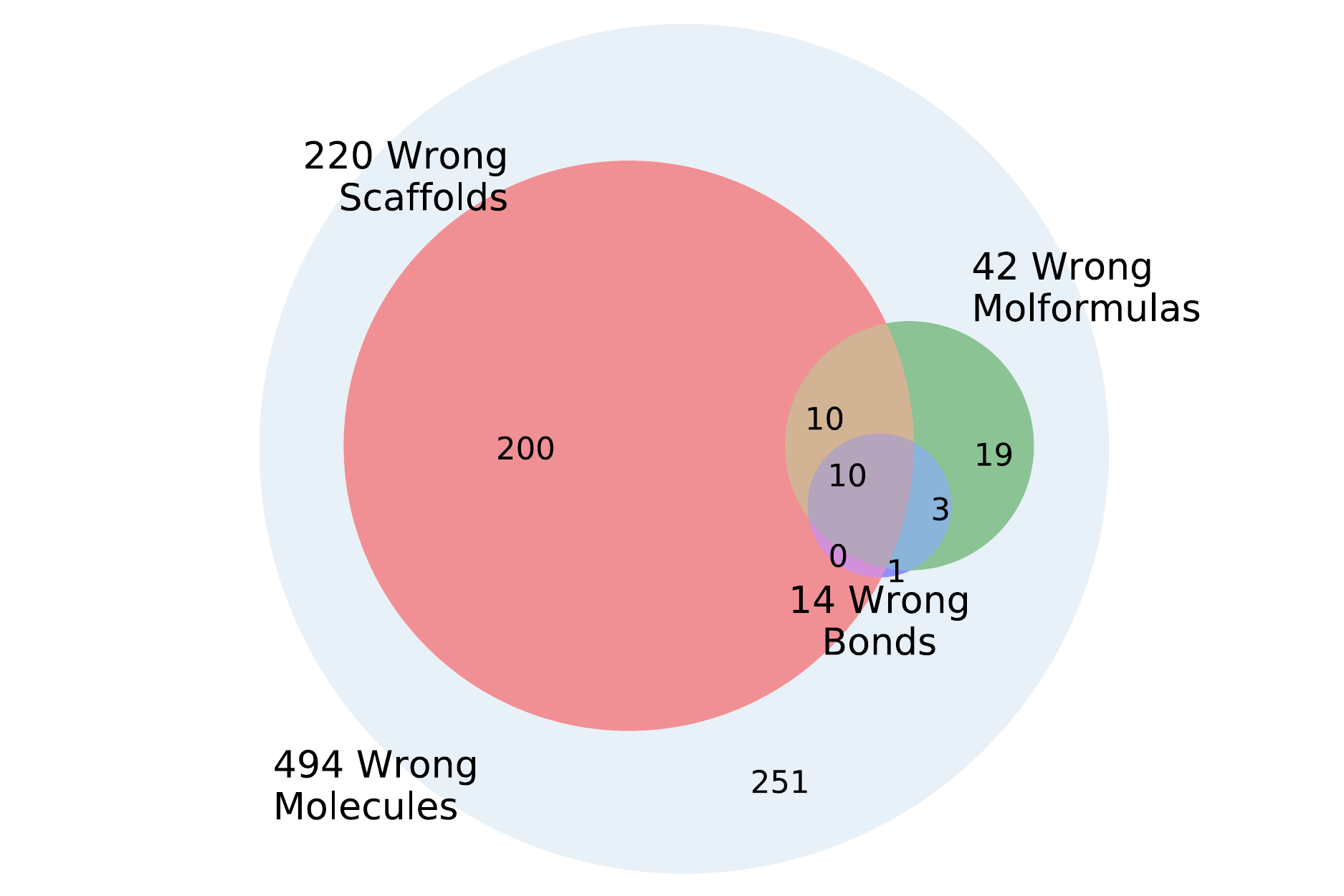}
\end{figure}

The models in general produce large percentages of valid SMILES (Table
\ref{tab:Model Properties}). However, using enumeration in the input
and output significantly increases the percentage of the outputs where
the decoded molecule is not the same as the encoded molecule. The
input and output molecules were further compared with regard to scaffold,
molecular sum formula and equality of bonds. Figure \ref{fig:Venn-diagram-of}
 shows the error types and overlap for the can2enum model. 494 molecules
out of 1000 tested was valid SMILES but not the same as the input
molecule. Further 220 had the wrong scaffold, 42 the wrong sum formula
and 14 the wrong bondtypes or number of bonds. The majority (251)
had the right scaffold, the right atoms and the correct bonds, but
had seemingly assembled the molecule in a wrong order. The bond types
and atoms are in principle simple accounting operations independant
of the SMILES enumeration, whereas the models struggle more with the
scaffold reconstruction and the atom order, which are influenced by
the SMILES enumeration. The results for the other heteroencoder models
are qualitatively similar (not shown).

\subsubsection*{Enumeration Challenge}

\begin{figure*}
\caption{\label{fig:SMILES-enumeration-challenge}SMILES enumeration challenge
of the GDB-8 dataset based Enum2can and Can2enum encoders. The same
three molecules were encoded from 10 enumerated SMILES and projected
to the latent space reduced to two dimensions with principal components
analysis (PCA). Using enumerated SMILES for training of the encoder
leads to the tightest clustering, but also training with the enumerated
SMILES in the decoder improves the clustering (c.f. Figure \ref{fig:Can2can-Enumeration-challenge}).
Small blue dots are the test set used for the PCA reduction.}

\includegraphics[width=1\textwidth]{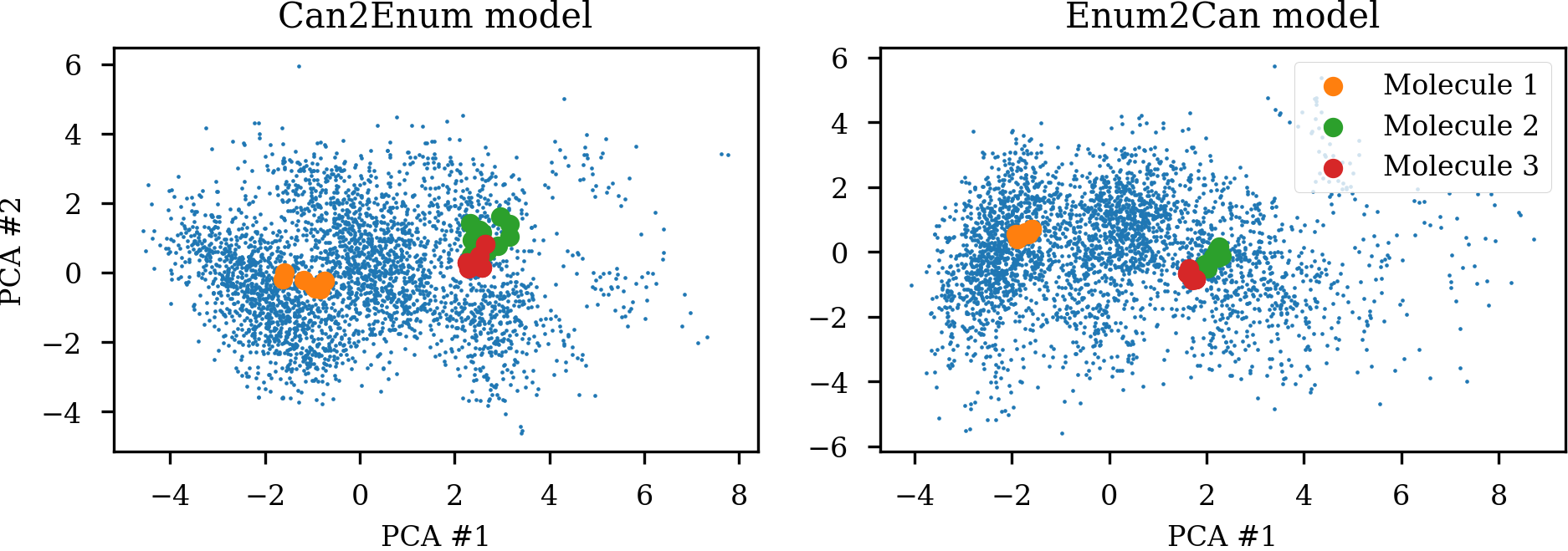}
\end{figure*}

The encoders capabilities to handle different SMILES from the same
molecule were tested by projection to a PCA reduction of the latent
space (c.f. Figure \ref{fig:Can2can-Enumeration-challenge}). Figure
\ref{fig:SMILES-enumeration-challenge}
 shows the improvement that can be obtained by training the heteroencoders
with enumerated SMILES for either the encoder and decoder. Training
the encoder with enumerated SMILES strings gives the tightest clustering
(enum2can), showing that the encoder has learned to recognize the
same molecule independent on actual serialization of the SMILES string.
By showing multiple different SMILES strings to the encoder during
training, the encoder can produce the latent space coordinate most
suitable for recreating the SMILES form of the decoder, irrespective
of the SMILES form shown to the encoder. The enum2enum model has a
similar tight clustering as the enum2can model (not shown). The can2enum
model also show more tight clustering than the can2can model from
Figure \ref{fig:Can2can-Enumeration-challenge}, indicating that the
heteroencoding itself changes the latent space although the encoder
itself was not trained on different SMILES forms. Alternatively, the
model is doing a more complicated task which could work as regularization
leading to better generalization.

\subsubsection*{Sampling using probability distribution}

\begin{figure*}
\caption{\label{fig:Multinomial-sampling}Multinomial sampling of the decoder
for two different models illustrated with heat maps of the character
probability for each step during decoding of the latent space. A:
The can2can model is very certain at each step and samples the same
canonical SMILES each time. B: The can2enum model has more possibilities
at each step in the beginning. The probability heatmap and sampled
SMILES will be different for each sampling run, depending on which
character is chosen from the probability distribution at each step.}

\includegraphics[width=1\textwidth]{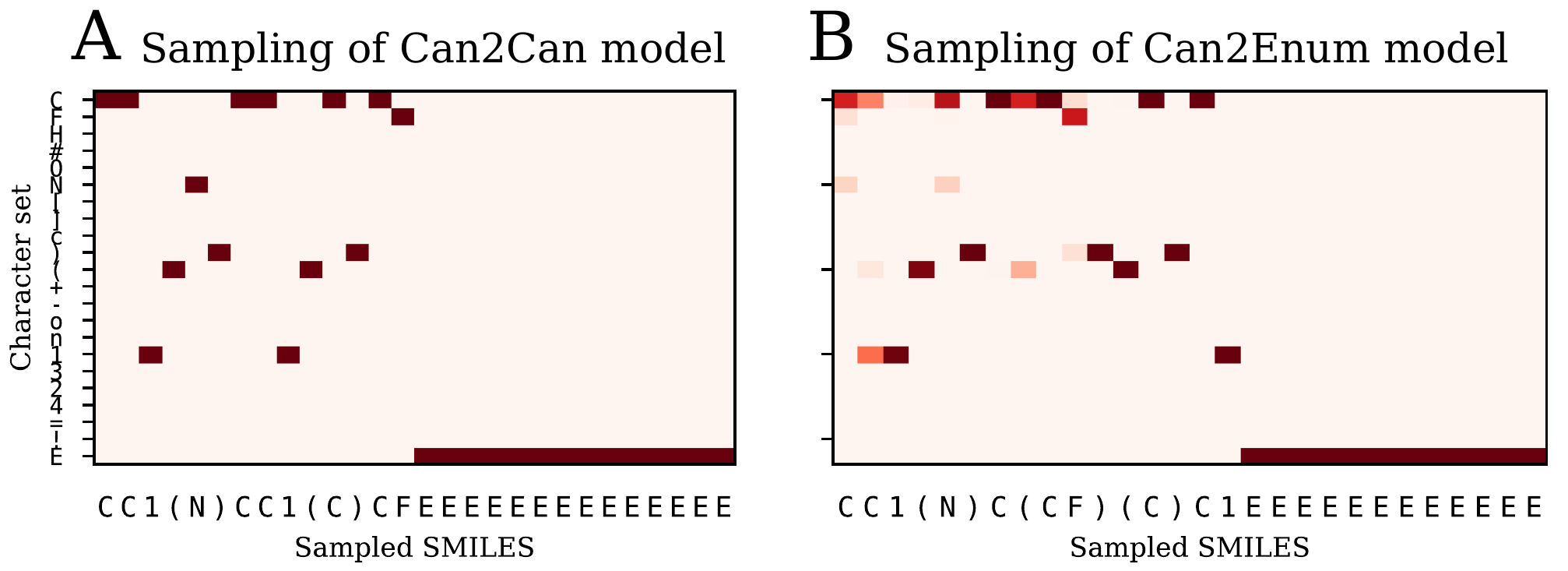}
\end{figure*}

\begin{figure}
\caption{\label{fig:Examples-of-sampled}Examples of different sampled molecules
using multinomial sampling with the decoder from the two layer LSTM
model (enum2enum 2-layer). The one in the upper-left corner is the
reference molecule used to encode the latent space coordinates.}

\includegraphics[width=1\columnwidth]{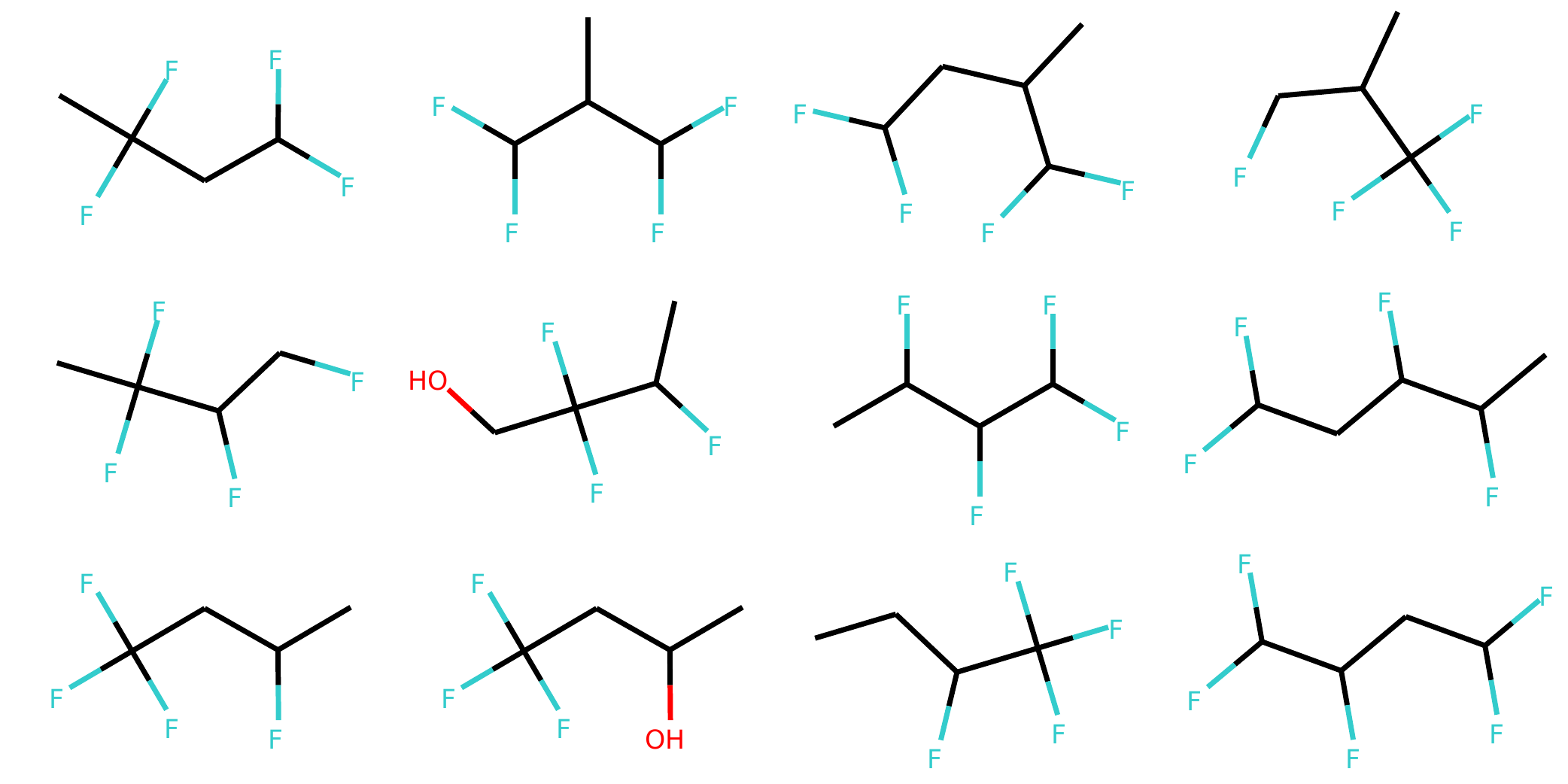}
\end{figure}

Figure \ref{fig:Multinomial-sampling}
 illustrates the difference between probability sampling of the can2can
and can2enum model. The decoder outputs a probability distribution
at each step, which can be sampled randomly according to the probabilities
(Multinomial sampling). For the can2can model, there is little difference
between this sampling strategy and the simple selection of the most
probable next character. The can2enum model instead show a lot more
uncertainty in the next characters in the beginning of the sampling.
The first character is most likely ``C'', but also ``N'' and ``F''
are possibilities. As the model samples ``C'', the next character
is either a ``C'', a branching ``(``, or start of a ring ``1''.
Because it then samples ``C'', it has to choose a ring start next.
Towards the end of sampling, the decoder gets completely certain with
the last 6 characters, probably because there is only one way to finish
the molecule with the already sampled characters. 
\begin{table*}
\caption{\label{tab:Stats_sampled}Statistics on molecule generation with multinomial
sampling at t=1.0, n=1000, GDB-8 dataset based models.}

\begin{tabular}{cccc}
\hline 
 & Can2Can & Can2Enum & Enum2Enum 2-Layer\tabularnewline
\hline 
\hline 
Unique SMILES & 1 & 315 & 111\tabularnewline
\% Correct Mol & 100 & 20 & 57\tabularnewline
Unique SMILES for correct Mol & 1 & 34 & 42\tabularnewline
Unique Molecules & 1 & 88 & 17\tabularnewline
Average Fingerprint Similarity & 1.0 & 0.27 & 0.32\tabularnewline
\hline 
\end{tabular}
\end{table*}
Table \ref{tab:Stats_sampled} shows some statistics on the sampled
molecules using the latent coordinates from a single molecule. The
model trained on canonical SMILES in both encoder and decoder are
very sure about the SMILES it want to recreate, as only one SMILES
form and one molecule is sampled. In contrast, the decoders trained
with the enumerated SMILES create different SMILES forms of the correct
molecule, but also creates other molecules as well. The more complex
model with two LSTM layers handles the task a bit better than the
single layer version, which only produce the molecule presented for
the encoder 20\% of the times. Examples of the sampled molecules from
the two layer model are shown in Figure \ref{fig:Examples-of-sampled}.

\subsection*{QSAR modelling using ChEMBL trained heteroencoders}

Training of the models used on the ChEMBL datasets, resulted in final
losses of approximately 0.001, 0.01, 0.10, 0.11 for the can2can, enum2can,
can2enum and enum2enum configurations of the training sets, respectively
Reconstruction performance of the different encoder/decoder configurations
is presented in the Table \ref{tab:CHEMBL_recon}.
\begin{table*}
\caption{\label{tab:CHEMBL_recon}Reconstruction performance on the ChEMBL
datasets of the different encoder/decoder configurations}

\begin{tabular}{cccc}
\hline 
ChEMBL model & Invalid SMILES (\%) & SMILES different from input (\%) & Wrong molecules (\%)\tabularnewline
\hline 
Can2Can & 0.2 & 0.3 & 0.1\tabularnewline
Enum2Can & 9.3 & 42.5 & 36.6\tabularnewline
Can2Enum & 9.3 & 99.9 & 65.6\tabularnewline
Enum2Enum & 6.7 & 100 & 69.9\tabularnewline
\hline 
\end{tabular}
\end{table*}
 The performance of the QSAR modelling done based on ECFP4 fingerprints
and the different vectors obtained from the bottleneck layers of the
three different neural network models are shown in Table \ref{tab:QSAR-performance}.
\begin{table*}
\caption{\label{tab:QSAR-performance}Performance of the QSAR models on the
held out test-set for different input data. The best performance for
each metric and dataset is highlighted in bold. R\textsuperscript{2}
is the squared correlation coefficient (closer to one is better),
RMSE is the root mean square error of prediction on the test set (lower
is better).}

\begin{tabular}{ccccccccccccc}
\hline 
 & \multicolumn{2}{c}{IGC50} & \multicolumn{2}{c}{LD50} & \multicolumn{2}{c}{BCF} & \multicolumn{2}{c}{Solubility} & \multicolumn{2}{c}{MP} & \multicolumn{2}{c}{Average}\tabularnewline
\cline{2-13} 
Input Type & R\textsuperscript{2} & RMSE & R\textsuperscript{2} & RMSE & R\textsuperscript{2} & RMSE & R\textsuperscript{2} & RMSE & R\textsuperscript{2} & RMSE & R\textsuperscript{2} & RMSE\textsuperscript{{*}}\tabularnewline
\hline 
Enum2Enum & \textbf{0.81} & \textbf{0.43} & \textbf{0.68} & \textbf{0.54} & 0.73 & 0.71 & \textbf{0.90} & \textbf{0.65} & 0.86 & \textbf{37} & \textbf{0.80} & \textbf{0.75}\tabularnewline
Can2Enum & 0.78 & 0.46 & \textbf{0.68} & \textbf{0.54} & \textbf{0.74} & \textbf{0.69} & 0.89 & 0.69 & 0.86 & \textbf{37} & 0.79 & 0.77\tabularnewline
Enum2Can & 0.78 & 0.46 & 0.65 & 0.57 & 0.73 & 0.71 & \textbf{0.90} & 0.66 & \textbf{0.87} & 38 & 0.78 & 0.78\tabularnewline
Can2Can & 0.71 & 0.53 & 0.59 & 0.62 & 0.66 & 0.79 & 0.82 & 0.87 & 0.82 & 43 & 0.72 & 0.89\tabularnewline
ECFP4 & 0.60 & 0.62 & 0.62 & 0.59 & 0.53 & 0.94 & 0.65 & 1.21 & 0.82 & 43 & 0.64 & 1.00\tabularnewline
\hline 
\multicolumn{13}{c}{\textsuperscript{}{*}) RMSE normalized using the RMSE of ECFP4 based
models before averaging}\tabularnewline
\end{tabular}
\end{table*}
 There is some improvement from the ECFP based baseline models to
the can2can vector based models, with further improvement for the
models based on the latent vectors extracted with the heteroencoders.
The three different heteroencoders seem to produce latent vectors
which perform very similar to each other in the QSAR modelling, with
a tendency for the average performance to rise from enum2can to enum2enum
over can2enum.

An interesting observation was that approximately 40\% of the neurons
of the bottleneck layer for each configuration are never activated.
This is likely related to the use of the ReLU activation function
and diminishing or enlarging the bottleneck layer resulted in almost
the same percentage of inactive neurons (results not shown). The latent
vector is thus even denser than the chosen number of neurons.

\section*{Discussion}

Changing the representations used for training autoencoders (here
called heteroencoders) have a marked influence on the properties and
organization of the latent space. Although a perfect correlation to
the standard fingerprint similarity is not wanted or expected, it
is more reassuring that the dependence to the SMILES sequences is
at a similar level to the fingerprint based similarity, than the situation
where the correlation to the SMILES sequence is much larger than the
correlation to the fingerprint metric. The greater balance between
the two correlations strongly indicates that the latent space is just
as relevant for the molecular scaffold as it is to the SMILES sequence
in itself. 

The dataset used in the first part of this study was of a limited
size and molecular complexity (only 8 atoms). Additionally, as the
dataset is fully enumerated the same graph structures are very likely
present in both training and test set, which could be the basis of
the excellent reconstructions of the test sets. The model could in
principle memorize all graph structures instead of learning the rules
behind the graph scaffolds, and then simply assign a specific sequence
of atoms to the memorized graph. Even though the dataset was somewhat
simple, the models trained on enumerated data may have struggled because
of low neural network fitting capacity. This indeed seems to be the
case as the 2-layer enum2enum model has much lower final loss (Table
\ref{tab:Model Properties}) and also much better reconstruction statistics
when reconstructing (Table \ref{tab:Model Properties}) and sampling
the molecules (Table \ref{tab:Stats_sampled}) than the single layer
enum2enum model. Even though there is a rough correlation between
the SMILES validity rate and the molecule reconstruction error rate,
with heteroencoders the molecule reconstruction rate becomes a more
relevant term to measure than the SMILES validity rate, as the former
can diverge a lot, while the SMILES validity error rate is still low.

Models employed in other studies are more complex with larger and
multiple layers\cite{Gomez-Bombarelli2016,goh2017chemception,Bjerrum2017_molgen,Blaschke2017_autoencoders,Li2018,Xu2017_seq2seq}.
The heteroencoder concept was thus further expanded to also handle
ChEMBL datasets. The expansion of the networks to two layers in both
the encoder and decoder, use of bidirectional layers in the encoder
and a larger number of LSTM cells allowed to fit the larger molecules,
although the uncertainty in the reconstruction of the molecule is
still present (c.f. Table \ref{tab:CHEMBL_recon}). It is likely that
even more complex architectures with three LSTM layers or a further
enlargement of the number of LSTM cells would be needed to lower the
molecule reconstruction error further.

The image to sequence model seems to be an outlier in comparison with
the SMILES based models, in the respect that the latent space don't
show much correlation with neither the SMILES sequence to be decoded
or the molecular graph. The model produces a very low percentage of
invalid SMILES and also has a low error rate with respect to molecule
reconstruction, but is also decoding to canonical SMILES, which is
an easier task then decoding to enumerated SMILES. The various other
tests showed no big difference or benefit when compared to the much
simpler use of different SMILES representations. On the other hand,
the heteroencoder architecture may be useful for architectural experiments
with large unlabeled datasets to find better architectures and suitable
deep learning feature extractions for training on 2D embeddings of
molecules. The identified architectures and trained weights may be
useful for transfer learning in for example QSAR modeling.

The failure of the image to SMILES heteroencoder to produce significantly
better latent representation fits with the observation that the latent
space is mostly influenced by the decoding procedure, not the encoding
procedure. The various encoders, whether based on images, canonical
SMILES or trained on enumerated SMILES, seem to learn to recognize
the molecules anyway and create a latent space that is best suitable
for recreation of the decoders form. It thus seems that using enumeration
techniques or other formats for the decoder will influence the latent
space the most. 

Training autoencoders on enumerated or different data further seems
to improve the latent space with respect to its relevance for QSAR
modelling. This is encouraging as it suggests that the extracted vectors
are not only relevant for reconstruction of the molecular scaffold
in itself, but additionally capture the variations underlying biological
as well as physico-chemical properties of the molecules. It seem that
already the encoder independence of the SMILES form for the enum2can
leads to a more smooth latent space (c.f. Figure \ref{fig:SMILES-enumeration-challenge}
panel A), which increases the relevance for QSAR modelling. This is
in contrast to the results in Table \ref{tab:Model Properties}, where
a less skewed correlation to the decoded SMILES serialization in the
encoder part is forced by training on enumerated data in the output,
which however only leads to marginal gains in QSAR model performance. 

The improvement seems quite marked and larger than what other studies
have found. Winter et al. also used the heteroencoder approach in
parallel to our work and found improvements for QSAR modelling\cite{Winter2018}.
However, the improvements over baseline models were not as marked
as in our results. The differences in network architectures (our use
of bidirectional layers, LSTM vs. GRU's and batch normalization as
example) and maybe also the choice of training data (Drug like molecules
of ChEMBL) could be possible explanations. Future benchmarking on
common datasets will likely show the way to the best network architecture
and what unlabelled datasets to use for specific tasks.

On the other hand, the solubility dataset we used have previously
been carefully modelled with chosen features and topological descriptors,
resulting in a R\textsuperscript{2} of 0.92 and a standard deviation
of prediction of the test set of 0.6\cite{huuskonen2000estimation}.
Likewise, a carefully crafted QSAR model of BCF obtained a R\textsuperscript{2}
of 0.73 and an RMSE of 0.69 \cite{gissi2014alternative}, which is
on par with our model using the can2enum derived latent vectors. However
a later benchmark showed better performance for the CORAL software
for prediction of BCF (R\textsuperscript{2}: 0.76, RMSE: 0.64)\cite{gissi2015},
suggesting that further improvements are possible.

Thus the QSAR models based on heteroencoder derived latent vectors
seem to almost match the performance of highly optimized QSAR models
from selected features (c.f. Table \ref{tab:QSAR-performance}), and
it may rather be the ECFP4 and can2can model derived latent vectors
that are mediocre for the tested type of QSAR tasks. Further, the
ECFP fingerprints and auto-/heteroencoder derived latent vectors are
of different dimension and nature. The fingerprints are 1024 dimensional,
but binary, where the latent vectors are 256 dimensional and real
valued. To make sure that the improvements were not due to different
optima of the model hyper parameters for the different data, the neural
network architectures for the QSAR models were optimized based on
the ECFP4 fingerprint input. Some improvement of the fingerprint based
models were observed, but reusing the ECFP4 hyper parameters for the
latent vector based modelling still resulted in a large improvement
in model performance for these input types. Further tuning of the
hyper parameters of the models based on the latent vectors could likely
further increase the performance to some degree (not tested). On the
other hand, the denser dimensionality (256 <\textcompwordmark{}< 1024)
could help protect against over fitting and make the choice of hyper
parameters less critical for these models. Either way, the use of
heteroencoder derived latent vectors seem to be the better choice.

Feature generation for a dataset of chemical structures using the
ChEMBL trained auto-/heteroencoders described in the publication is
publicly available and hosted on the OSDR platform\cite{OSDR_features2018},
where it is possible to encode molecular datasets into the latent
vector space for subsequent uses, such as in QSAR modelling.

The increased relevance of the latent space with respect to bioactivity
and physico-chemical properties are likely to increase the relevance
and quality of the \textit{de-novo} generated libraries where the
neighborhood of as example lead compounds are sampled on purpose.
However, the use of enumeration for training the decoder comes at
the cost of greater uncertainty in the decoding, at a marginal improvement
to the relevance of the latent space for QSAR modelling when compared
to the enum2can model. On the other hand, the greater uncertainty
and ``creativity'' in decoding could be beneficial and further help
in creating more diversity in the generated libraries, but if this
is the case has yet to be investigated. The choice of enumeration
for decoder and/or encoder will thus likely depend on the intended
use-cases.

\section*{Conclusion}

The pilot study using a fully enumerated train and test set with 8
atoms showed that the latent space representation is sensitive to
the chosen formats of the input and output in the training. Using
canonical SMILES for the decoder gives a latent space representation
which seem closer correlated to the SMILES strings than the molecular
graphs. In contrast, training the encoders on input and output from
different representations in chemical heteroencoders (image or enumerated
non-canonical SMILES), gives a latent space with a better balance
between SMILES similarity and a traditional molecular similarity metric.
Forcing the encoder-decoder pair to trans-code between different formats
or SMILES representation indeed seem to give a latent space closer
to an abstract idea or encoding of the underlying molecule. The latent
space properties were mostly found to be influenced by the choice
of training data and representations used as the decoder targets.
The multinomial sampling of the decoder shows higher variance for
the decoders trained to predict enumerated data, both with regard
to the SMILES form of the correct molecule but also by sampling other
molecules. The changed properties of the decoders have broken the
dependence on producing canonical SMILES, and may make them more relevant
in \textit{de-novo} design approaches in drug discovery, where a balance
between similarity and variance is the goal.

The improved performance when using the latent space vectors from
heteroencoders for QSAR modelling, further emphasizes their increased
relevance, not just being a more SMILES independent representation
of the molecule, but also for a better description of the chemical
space relevant for biological as well as physico-chemical properties.
This should hopefully lead to more drug-discovery relevant de-novo
generated libraries. The increased relevance however comes at the
price of greater uncertainty in the decoding, although more complex
decoders seem to perform better at that metric.

\section*{Contributions}

Esben Jannik Bjerrum developed the concept of heteroencoders, performed
the training and tests using the GDB-8 datasets and prepared the initial
\cite{huuskonen2000estimation} and final manuscripts. Boris Sattarov
optimized the network architecture for ChEMBL data, prepared the QSAR
datasets, trained and tested the performance of the latent vectors
for QSAR application and helped prepare the final manuscript.

\section*{Acknowledgemets}

We thank Dr. Alexandru Korotcov, science team leader of ScienceDataSoftware,
for helpful comments on the manuscript.

\section*{Conflict of interests}

E. J. Bjerrum is the owner of Wildcard Pharmaceutical Consulting.
The company is usually contracted by biotechnology/pharmaceutical
companies to provide third party contract research and IT services.
Boris Sattarov is employed by Science Data Software LLC on a part
time basis. The company provide data infrastructure and machine learning
capabilities for drug discovery and chemical research.

\bibliographystyle{elsarticle-num}
\bibliography{HetEncoders_db}

\begin{thebibliography}{10}
\expandafter\ifx\csname url\endcsname\relax
  \def\url#1{\texttt{#1}}\fi
\expandafter\ifx\csname urlprefix\endcsname\relax\def\urlprefix{URL }\fi
\expandafter\ifx\csname href\endcsname\relax
  \def\href#1#2{#2} \def\path#1{#1}\fi

\bibitem{Gomez-Bombarelli2016}
R.~G{\'o}mez-Bombarelli, D.~Duvenaud, J.~M. Hern{\'a}ndez-Lobato,
  J.~Aguilera-Iparraguirre, T.~D. Hirzel, R.~P. Adams, A.~Aspuru-Guzik,
  Automatic chemical design using a data-driven continuous representation of
  molecules, arXiv preprint arXiv:1610.02415.

\bibitem{Blaschke2017_autoencoders}
T.~Blaschke, M.~Olivecrona, O.~Engkvist, J.~Bajorath, H.~Chen, Application of
  generative autoencoder in de novo molecular design, Molecular Informatics
  37~(1-2)  1700123.
\newblock \href {http://dx.doi.org/10.1002/minf.201700123}
  {\path{doi:10.1002/minf.201700123}}.

\bibitem{Xu2017_seq2seq}
Z.~Xu, S.~Wang, F.~Zhu, J.~Huang, Seq2seq fingerprint: An unsupervised deep
  molecular embedding for drug discovery, in: Proceedings of the 8th ACM
  International Conference on Bioinformatics, Computational Biology,and Health
  Informatics, ACM-BCB '17, ACM, New York, NY, USA, 2017, pp. 285--294.
\newblock \href {http://dx.doi.org/10.1145/3107411.3107424}
  {\path{doi:10.1145/3107411.3107424}}.

\bibitem{Chung2014}
J.~Chung, C.~Gulcehre, K.~Cho, Y.~Bengio, Empirical evaluation of gated
  recurrent neural networks on sequence modeling, arXiv preprint
  arXiv:1412.3555.

\bibitem{Hochreiter1997}
S.~Hochreiter, J.~Schmidhuber, Long short-term memory, Neural computation 9~(8)
  (1997) 1735--1780.

\bibitem{Weininger1970}
D.~Weininger, {SMILES}, a chemical language and information system. 1.
  introduction to methodology and encoding rules, in: Proc. Edinburgh Math.
  SOC, Vol.~17, 1970, pp. 1--14.

\bibitem{Bjerrum2017}
E.~J. Bjerrum, Smiles enumeration as data augmentation for neural network
  modeling of molecules, arXiv preprint arXiv:1703.07076.

\bibitem{Li2018}
Y.~Li, L.~Zhang, Z.~Liu, Multi-objective de novo drug design with conditional
  graph generative model, arXiv preprint arXiv:1801.07299.

\bibitem{goh2017chemception}
G.~B. Goh, C.~Siegel, A.~Vishnu, N.~O. Hodas, N.~Baker, Chemception: A deep
  neural network with minimal chemistry knowledge matches the performance of
  expert-developed qsar/qspr models, arXiv preprint arXiv:1706.06689.

\bibitem{Polishchuk2013}
P.~G. Polishchuk, T.~I. Madzhidov, A.~Varnek, Estimation of the size of
  drug-like chemical space based on gdb-17 data., Journal of computer-aided
  molecular design 27 (2013) 675--679.
\newblock \href {http://dx.doi.org/10.1007/s10822-013-9672-4}
  {\path{doi:10.1007/s10822-013-9672-4}}.

\bibitem{Ruddigkeit2012}
L.~Ruddigkeit, R.~Van~Deursen, L.~C. Blum, J.-L. Reymond, Enumeration of 166
  billion organic small molecules in the chemical universe database gdb-17,
  Journal of chemical information and modeling 52~(11) (2012) 2864--2875.

\bibitem{Gaulton2017}
A.~Gaulton, A.~Hersey, M.~Nowotka, A.~P. Bento, J.~Chambers, D.~Mendez,
  P.~Mutowo, F.~Atkinson, L.~J. Bellis, E.~Cibrián-Uhalte, M.~Davies,
  N.~Dedman, A.~Karlsson, M.~P. Magariños, J.~P. Overington, G.~Papadatos,
  I.~Smit, A.~R. Leach, The chembl database in 2017, Nucleic Acids Research
  45~(D1) (2017) D945--D954.
\newblock \href {http://dx.doi.org/10.1093/nar/gkw1074}
  {\path{doi:10.1093/nar/gkw1074}}.

\bibitem{EpiSuiteDownload}
\href{http://esc.syrres.com/interkow/EpiSuiteData.htm}{Epi suite data}, Online
  (Jul. 2018).
\newline\urlprefix\url{http://esc.syrres.com/interkow/EpiSuiteData.htm}

\bibitem{EpiSuiteSoftware}
U.~EPA, Estimation programs interface suite™ for microsoft® windows, v 4.11,
  United States Environmental Protection Agency, Washington, DC, USA. (2018).

\bibitem{huuskonen2000estimation}
J.~Huuskonen, Estimation of aqueous solubility for a diverse set of organic
  compounds based on molecular topology, Journal of Chemical Information and
  Computer Sciences 40~(3) (2000) 773--777.

\bibitem{Arnot2006}
J.~A. Arnot, F.~A. Gobas, A review of bioconcentration factor (bcf) and
  bioaccumulation factor (baf) assessments for organic chemicals in aquatic
  organisms, Environmental Reviews 14~(4) (2006) 257--297.
\newblock \href {http://dx.doi.org/10.1139/a06-005}
  {\path{doi:10.1139/a06-005}}.

\bibitem{Schultz1997}
T.~W. Schultz, Tetratox: Tetrahymena pyriformis population growth impairment
  endpointa surrogate for fish lethality, Toxicology Methods 7~(4) (1997)
  289--309.
\newblock \href {http://dx.doi.org/10.1080/105172397243079}
  {\path{doi:10.1080/105172397243079}}.

\bibitem{ChemIDplus}
\href{http://chem.sis.nlm.nih.gov/chemidplus/chemidheavy.jsp}{Chemidplus
  database}, Online (Apr. 2018).
\newline\urlprefix\url{http://chem.sis.nlm.nih.gov/chemidplus/chemidheavy.jsp}

\bibitem{mendeleev2014}
L.~Mentel, \href{https://bitbucket.org/lukaszmentel/mendeleev}{{mendeleev} -- a
  python resource for properties of chemical elements, ions and isotopes}.
\newline\urlprefix\url{https://bitbucket.org/lukaszmentel/mendeleev}

\bibitem{scikit-learn}
F.~Pedregosa, G.~Varoquaux, A.~Gramfort, V.~Michel, B.~Thirion, O.~Grisel,
  M.~Blondel, P.~Prettenhofer, R.~Weiss, V.~Dubourg, J.~Vanderplas, A.~Passos,
  D.~Cournapeau, M.~Brucher, M.~Perrot, E.~Duchesnay, Scikit-learn: Machine
  learning in {P}ython, Journal of Machine Learning Research 12 (2011)
  2825--2830.

\bibitem{Landrum2016}
G.~A. Landrum, \href{http://www.rdkit.org/,
  https://github.com/rdkit/rdkit}{Rdkit: Open-source cheminformatics software}
  (2016).
\newline\urlprefix\url{http://www.rdkit.org/, https://github.com/rdkit/rdkit}

\bibitem{walt2011numpy}
S.~v.~d. Walt, S.~C. Colbert, G.~Varoquaux, The numpy array: a structure for
  efficient numerical computation, Computing in Science \& Engineering 13~(2)
  (2011) 22--30.

\bibitem{chollet2015}
F.~Chollet, \href{https://github.com/fchollet/keras}{keras} (2015).
\newline\urlprefix\url{https://github.com/fchollet/keras}

\bibitem{Abadi2016b}
M.~Abadi, A.~Agarwal, P.~Barham, E.~Brevdo, Z.~Chen, C.~Citro, G.~Corrado,
  A.~Davis, J.~Dean, M.~Devin, et~al., Tensorflow: Large-scale machine learning
  on heterogeneous distributed systems. 2016, arXiv preprint arXiv:1603.04467.

\bibitem{Nair2010}
V.~Nair, G.~E. Hinton, Rectified linear units improve restricted boltzmann
  machines, in: Proceedings of the 27th international conference on machine
  learning (ICML-10), 2010, pp. 807--814.

\bibitem{Williams1989}
R.~J. Williams, D.~Zipser, A learning algorithm for continually running fully
  recurrent neural networks 1 (1989) 270--280.
\newblock \href {http://dx.doi.org/10.1162/neco.1989.1.2.270}
  {\path{doi:10.1162/neco.1989.1.2.270}}.

\bibitem{Kingma2014}
D.~Kingma, J.~Ba, Adam: A method for stochastic optimization, arXiv preprint
  arXiv:1412.6980.

\bibitem{Szegedy2014}
C.~Szegedy, W.~Liu, Y.~Jia, P.~Sermanet, S.~Reed, D.~Anguelov, D.~Erhan,
  V.~Vanhoucke, A.~Rabinovich, Going deeper with convolutions\href
  {http://arxiv.org/abs/http://arxiv.org/abs/1409.4842v1}
  {\path{arXiv:http://arxiv.org/abs/1409.4842v1}}.

\bibitem{Cock2009_biopython}
P.~J.~A. Cock, T.~Antao, J.~T. Chang, B.~A. Chapman, C.~J. Cox, A.~Dalke,
  I.~Friedberg, T.~Hamelryck, F.~Kauff, B.~Wilczynski, M.~J.~L. de~Hoon,
  Biopython: freely available python tools for computational molecular biology
  and bioinformatics, Bioinformatics 25~(11) (2009) 1422--1423.
\newblock \href {http://dx.doi.org/10.1093/bioinformatics/btp163}
  {\path{doi:10.1093/bioinformatics/btp163}}.

\bibitem{Bemis1996}
G.~W. Bemis, M.~A. Murcko, The properties of known drugs. 1. molecular
  frameworks, Journal of Medicinal Chemistry 39~(15) (1996) 2887--2893, pMID:
  8709122.
\newblock \href {http://dx.doi.org/10.1021/jm9602928}
  {\path{doi:10.1021/jm9602928}}.

\bibitem{Bjerrum2017_molgen}
E.~J. Bjerrum, R.~Threlfall, Molecular generation with recurrent neural
  networks (rnns)\href {http://arxiv.org/abs/arXiv:1705.04612}
  {\path{arXiv:arXiv:1705.04612}}.

\bibitem{van1995python}
G.~Van~Rossum, F.~L. Drake~Jr, Python reference manual, Centrum voor Wiskunde
  en Informatica Amsterdam, 1995.

\bibitem{OpenDataRepository}
\href{http://osdr.dataledger.io/}{Open science data repository}, Science Data
  Software LLC, 14914 Bradwill Court, Rockville, Maryland 20850, United States
  (Jun. 2018).
\newline\urlprefix\url{http://osdr.dataledger.io/}

\bibitem{Bergstra2011}
R.~B. Y.~B. Bergstra, James~S., B.~Kégl., Algorithms for hyper-parameter
  optimization., Advances in neural information processing systems.

\bibitem{Hyperopt}
\href{https://github.com/hyperopt/hyperopt}{Hyperopt: Distributed asynchronous
  hyper-parameter optimization}, online (Jun. 2018).
\newline\urlprefix\url{https://github.com/hyperopt/hyperopt}

\bibitem{Winter2018}
R.~Winter, F.~Montanari, F.~No{\'e}, D.-A. Clevert, Learning continuous and
  data-driven molecular descriptors by translating equivalent chemical
  representations, chemrxiv.org\href
  {http://dx.doi.org/10.26434/chemrxiv.6871628.v1}
  {\path{doi:10.26434/chemrxiv.6871628.v1}}.

\bibitem{gissi2014alternative}
A.~Gissi, D.~Gadaleta, M.~Floris, S.~Olla, A.~Carotti, E.~Novellino,
  E.~Benfenati, O.~Nicolotti, An alternative qsar-based approach for predicting
  the bioconcentration factor for regulatory purposes, ALTEX-Alternatives to
  animal experimentation 31~(1) (2014) 23--36.

\bibitem{gissi2015}
A.~Gissi, A.~Lombardo, A.~Roncaglioni, D.~Gadaleta, G.~F. Mangiatordi,
  O.~Nicolotti, E.~Benfenati, Evaluation and comparison of benchmark qsar
  models to predict a relevant reach endpoint: The bioconcentration factor
  (bcf), Environmental research 137 (2015) 398--409.

\bibitem{OSDR_features2018}
{Open Science Data Repository},
  \href{http://ssp.dataledger.io/features}{Features computation beta} (Sep.
  2018).
\newline\urlprefix\url{http://ssp.dataledger.io/features}

\end{thebibliography}

\end{document}